# UTTERANCE PARTITIONING FOR SPEAKER RECOGNITION: AN EXPERIMENTAL REVIEW AND ANALYSIS WITH NEW FINDINGS UNDER GMM-SVM FRAMEWORK

Nirmalya Sen[a], Md Sahidullah[b, *], Hemant Patil[c], Shyamal Kumar Das Mandal[d], Krothapalli Sreenivasa Rao[e], Tapan Kumar Basu[f]

[a]R. H. Sapat College of Engineering Management Studies & Research, Nashik- 422005, India.

[b]Université de Lorraine, CNRS, Inria, LORIA, F-54000 Nancy, France.

[c]Dhirubhai Ambani Institute of Information and Communication Technology (DA-IICT), Gandhinagar- 382 007, India.

[d]Centre for Education Technology, Indian Institute of Technology Kharagpur, Kharagpur- 721302, India.

[e]Department of Computer science and Engineering, Indian Institute of Technology Kharagpur, Kharagpur- 721302, India.

[f]Academy of Technology, Adisaptagram, Hooghly, West Bengal-712121, India.

nirmalyasen75@gmail.com, md.sahidullah@inria.fr, hemant_patil@daiict.ac.in, sdasmandal@cet.iitkgp.ac.in, ksrao@iitkgp.ac.in, basutk06@rediffmail.com

[*]Corresponding Author

**Abstract:**

The performance of speaker recognition system is highly dependent on the amount of speech used in enrollment and test. This work presents a detailed experimental review and analysis of the GMM-SVM based speaker recognition system in presence of duration variability. This article also reports a comparison of the performance of GMM-SVM classifier with its precursor technique Gaussian mixture model- universal background model (GMM-UBM) classifier in presence of duration variability. The goal of this research work is not to propose a new algorithm for improving speaker recognition performance in presence of duration variability. However, the main focus of this work is on utterance partitioning (UP), a commonly used strategy to compensate the duration variability issue. We have analysed in detailed the impact of training utterance partitioning in speaker recognition performance under GMM-SVM framework. We further investigate the reason why the utterance partitioning is important for boosting speaker recognition performance. We have also shown in which case the utterance partitioning could be useful and where not. Our study has revealed that utterance partitioning does not reduce the data imbalance problem of the GMM-SVM classifier as claimed in earlier study. Apart from these, we also discuss issues related to the impact of parameters such as number of Gaussians, supervector length, amount of splitting required for obtaining better performance in short and long duration test conditions from speech duration perspective. We have performed the experiments with telephone speech from POLYCOST corpus consisting of 130 speakers.

# 1    INTRODUCTION

Human voice contains information related to the identity of the speaker. Speaker recognition is the technology of recognizing speakers from their voices [Kinnunen, T. and Li, H., 2010, Hansen, J.H. and Hasan, T., 2015]. Speaker recognition technology has various real-world applications where person identification or verification task is required. For example, it can be used for identifying the speaker of a speech segment assuming the speaker belongs to within a set of known speakers. On the other hand, this technique can also be used for determining whether two voice samples are spoken by the same person or not. A typical speaker recognition system has two main modules: a feature extraction unit or front-end and a classifier or back-end. The feature extraction step represents the raw speech signal in a compact form. The classifier creates the speaker models from the feature vectors. During the test, the features computed from the test speech and the models of the target speaker model are used for computing the similarity measure.

The speaker recognition system uses short-term features such as mel-frequency cepstral coefficients (MFCCs) or perceptual linear prediction (PLP) features as front-end [Kinnunen, T. and Li, H., 2010]. As a classifier, Gaussian mixture model- universal background model (GMM-UBM), GMM support vector machine (GMM-SVM), joint factor analysis (JFA), total variability, deep neural networks (DNN) are widely used [Poddar, A., Sahidullah, M. and Saha, G., 2017]. In recent years, fixed-dimensional representations of speech utterance are predominantly used for speaker representation modeling. One of the earliest works in this direction is GMM-SVM approach where the speech utterances are represented by concatenated means of the maximum-a-posteriori (MAP) adapted GMM called as supervector [Campbell, W.M., Sturim, D.E. and Reynolds, D.A., 2006]. Subsequently, i-vector method is developed which projects the high-dimensional GMM-supervector into low-dimensional total variability space [Dehak, N., Kenny, P.J., Dehak, R., Dumouchel, P. and Ouellet, P., 2010]. The state-of-the-art speaker recognition system rely on x-vector speaker embedding which is computed from the neural network bottleneck layers after training DNN in a speaker discriminative manner with MFCCs as input [Snyder, D., Garcia-Romero, D., Sell, G., Povey, D. and Khudanpur, S., 2018, April]. The speaker embeddings are further used with probabilistic linear discriminative analysis (PLDA) for similarity measure [Matějka, P., Glembek, O., Castaldo, F., Alam, M.J., Plchot, O., Kenny, P., Burget, L. and Černocky, J., 2011, May].

Speaker recognition system gives reasonably good recognition accuracy when adequate amount of speech data for enrollment and test are available from controlled conditions. However, in a practical situation, the recognition performance severely degrades due to the limited amount of data and the mismatch due to the varying environmental conditions, channel variability, intrinsic variabilities in human voice [Hautamäki, R.G., Sahidullah, M., Hautamäki, V. and Kinnunen, T., 2017], etc. In speaker recognition literature, various methods are proposed to improve the performance in presence of those practical challenges [Hansen, J.H. and Hasan, T., 2015].

The work is related to the limited data problem issue; however, the goal of this work is not to propose a new algorithm for improving speaker recognition performance in case of limited data. In contrast, we perform a detailed investigation of utterance partitioning where a longer speech utterance is divided into multiple short segments before speaker modeling with fixed-dimensional representation [Dehak, N., Kenny, P.J., Dehak, R., Dumouchel, P. and Ouellet, P., 2010]. The utterance partitioning is frequently used for speaker recognition with limited data. For example, utterance partitioning is used to compensate the duration mismatch in speaker recognition [Sen, N., 2014, Kanagasundaram, A., Dean, D., Sridharan, S., Ghaemmaghami, H. and Fookes, C., 2017, Kanagasundaram, A., Dean, D., Sridharan, S., Gonzalez-Dominguez, J., Gonzalez-Rodriguez, J. and Ramos, D., 2014]. Utterance partitioning is also used to address the data imbalance problem in GMM-SVM framework [Mak, M.W. and Rao, W., 2011]. Similar idea is also used for boosting speaker recognition performance with i-vectors [Rao, W. and Mak, M.W., 2013]. The latest x-vector [Snyder, D., Garcia-Romero, D., Sell, G., Povey, D. and Khudanpur, S., 2018, April] system also trains the neural network with smaller speech segments created from the long speech file as available with the original dataset. Even though the utterance partitioning helps in several ways in modern speaker recognition, in most cases the usage is ad-hoc and comprehensive understanding of the utterance partitioning is not present in speaker recognition literature. In this article, we have performed a Comprehensive experimental review and insightful analysis of utterance partitioning technique for speaker recognition context under GMM-SVM framework.

For this analysis, we have selected classical GMM-UBM system and more advanced GMM-SVM system. Both these systems are simple and based on a common framework involving UBM. Unlike the highly non-linear processing used in DNN, both the systems under consideration use linear mixture model and linear kernel which in turn helps in analyzing the system and interpreting the results.

The main contributions of this work can be summarized as follows:

- We have shown the effect of test speech duration for GMM-UBM and GMM-SVM system.
- We have shown that, for short duration test segments the recognition performance of GMM-SVM classifier is very poor and this poor performance is not due to the data imbalance problem of GMM-SVM classifier. In contradiction to the previous studies, we have demonstrated that, application of the utterance partitioning approach does not reduce the data imbalance problem of GMM-SVM classifier.

- We discuss in detail why GMM-UBM and GMM-SVM behave differently for short and long test speech.
- We have shown that the speaker recognition performance for GMM-SVM system for short test speech can be improved substantially with training utterance partitioning. However, the improvement is marginal for long test speech.
- We also comment on the choice of parameters such as number of Gaussians, supervector length, amount of splitting required for obtaining better performance in short and long duration test conditions.

The remainder of the paper is organized as follows. In Section 2 and 3, we have shown the experimental framework. In Section 4, we have compared the recognition performance of GMM-UBM classifier with GMM-SVM classifier for recognition of long and short duration test segments. In Section 5, we have investigated the effects of partitioning of training utterance on the performance of GMM-SVM classifier for recognition of long and short duration test segments. In Section 6, we have shown the effect of the dimension of the GMM supervector on the overlap between various classes in the GMM supervector domain. In Section 7, we have investigated the average number of support vectors required by positive class and negative class of GMM-SVM classifier before and after partitioning of training utterance.

## 2 TRAINING OF GMM-SVM CLASSIFIER IN UTTERANCE PARTITIONING FRAMEWORK

In this section, we provide a concise discussion on training utterance partitioning (UP) approach for speaker recognition under GMM-SVM classifier.

Let us consider an $N$ class problem. In the usual training process (i.e., training of GMM-SVM classifier without partitioning of training utterance approach), from the training speech of the hypothesized speaker, we extract the sequence, $\mathbf{X} = \{\mathbf{x}_1, \mathbf{x}_2, \cdots, \mathbf{x}_T\}$ of feature vectors. We apply all these feature vectors simultaneously to adapt the UBM and prepare the training GMM supervector for the hypothesized speaker. Hence, in the usual training process, from training utterance of each speaker we prepare only one training GMM supervector. As it is an *N* class problem, we must create *N* numbers of SVM models. For training of SVM classifiers, we have employed one against the rest approach. Therefore, in the usual training process, the positive class (i.e., genuine speaker class) will have only one training sample (i.e., one GMM supervector) and the negative class (i.e., impostor class) will have $N-1$ training samples (i.e., *N*-1 GMM supervectors). However, in case of training of GMM-SVM classifier with partitioning of training utterance approach we proceed as below:

From the training utterance of the speaker, we calculate the sequence $\mathbf{X}=\{\mathbf{x}_1,\mathbf{x}_2,\cdots,\mathbf{x}_T\}$ of feature vectors and divide the sequence into *P* numbers of subsequences as follows:

$$\mathbf{X}=\{\underbrace{\mathbf{x}_1,\mathbf{x}_2,\cdots,\mathbf{x}_K}_{\text{subsequence 1}},\underbrace{\mathbf{x}_{K+1},\mathbf{x}_{K+2},\cdots,\mathbf{x}_{2K}}_{\text{subsequence 2}},\cdots,\underbrace{\mathbf{x}_{T-K+1},\mathbf{x}_{T-K+2},\cdots,\mathbf{x}_T}_{\text{subsequence }P}\} \quad (1)$$

We adapt the UBM by using each subsequence separately and prepare one training GMM supervector from each subsequence. Therefore, from *P* numbers of subsequences, we prepare *P* numbers of training GMM supervectors for each speaker. At the time of training of SVM classifier for any speaker, we use all *P* numbers of supervectors from that speaker as the positive class data and all the $(N-1)P$ numbers of supervectors from the remaining $N-1$ classes as the negative class data. The block diagram for training of GMM-SVM classifier with partitioning of training utterance (UP) approach is given in Fig. 1.

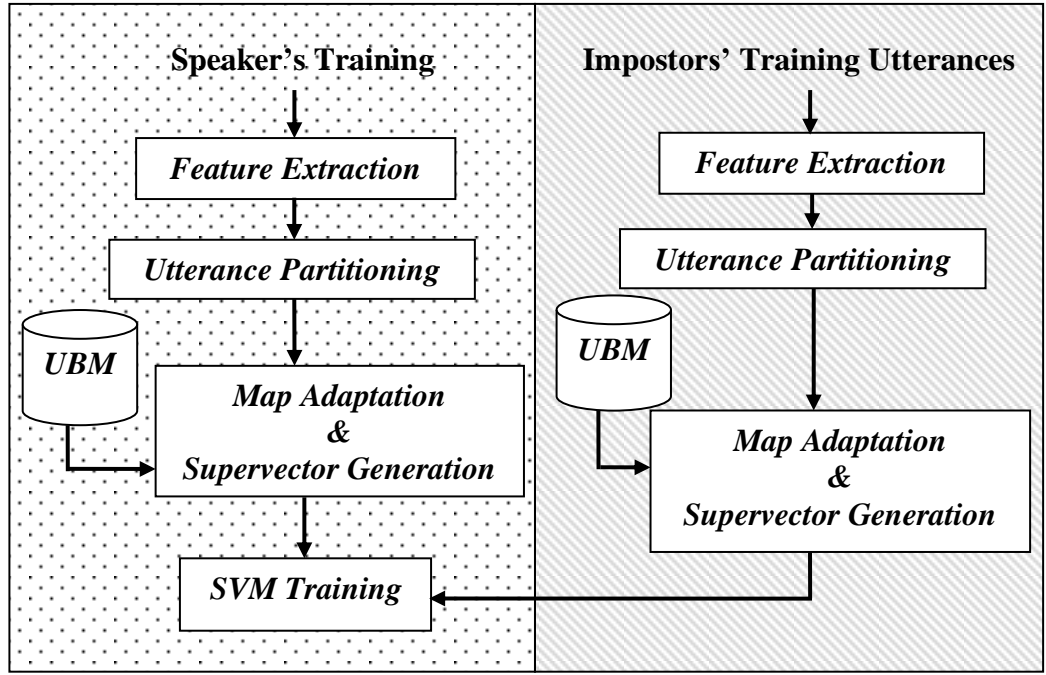

Fig. 1 Training of GMM-SVM classifier with partitioning of training utterance (UP) approach.

## 3     EXPERIMENTAL SETUPS

We have used the POLYCOST database [Petrovska, D., et al., 1998] for experimental evaluation of the performances of GMM-UBM classifier and GMM-SVM classifier (before application of the training utterance partitioning approach, as well as, after application of the training utterance partitioning approach). It is evident from literatures that MFCC is the most widely accepted feature set for speaker recognition application [Davis, S.B., Mermelsteine, P., 1980; Sahidullah, Md., Saha, G., 2012; Kinnunen T., 2004; Patil, H. A., 2005; Chakroborty, S., 2008; Kandali, A. B., 2012; Sahidullah, Md., 2015]. Therefore, we have also used the MFCC feature set for comparative evaluation of the two pattern classifiers. While computing the MFCC feature, we have performed the STFT operation on frame of size 20 milliseconds. The overlap between two consecutive frames was 50 %. We have used 20 Mel filters. After the application of DCT (i.e., discrete cosine transform), the first coefficient of the MFCC vector (i.e., DC value) is discarded since it contains only the energy of the spectrum and the remaining nineteen-dimensional MFCC feature vector is used.

## 3.1 Database Description

Majority of the studies in speaker recognition literature use NIST speaker recognition evaluation (SRE) datasets for evaluation and bench-marking. Those datasets are useful for studying speaker recognition performance in presence of variabilities such as acoustic environment, channel characteristics, spoken language, vocal effort, etc. Different data-driven variability compensation methods such as within-class co-variance normalization (WCCN) or nuisance attribute projection (NAP) are subsequently developed to remove the unwanted channel and environmental effects from speech representations [Dehak, N., Kenny, P.J., Dehak, R., Dumouchel, P. and Ouellet, P., 2010]. The careful selection of audio data for training WCCN and NAP parameters is also an important issue when applying those compensation methods. On the other hand, our goal in this work is rather different and is confined to the analysis of utterance partitioning and its impact on speaker recognition. In order to avoid the use of data-driven channel and environmental compensation methods, we choose a relatively homogeneous corpus POLYCOST having negligible channel variability.

The POLYCOST database was recorded as a common initiative within the COST 250 action during January-March 1996 [Petrovska, D., et al., 1998]. It contains around 10 sessions recorded by 134 speakers from 14 countries. Each session consists of 14 items. The database was collected through the European telephone network. The recording has been performed with ISDN cards on two XTL SUN platforms with an 8 kHz sampling frequency. Four speakers (M042, M045, M058 and F035) are not included in our experiments as each of them provide less than six sessions. All speakers (i.e., 130 after deletion of 4 from a total of 134 speakers) in the database were registered as clients. We have used the speech data of first five sessions to train the classifier (i.e., to prepare the speaker model), and used all the remaining speech data (i.e., from session six to last available session for respective speaker) for testing.

## 3.2 Preparation of GMM-UBM Classifier

We have created a single gender-independent universal background model (UBM). From the 130 speakers (as we explained earlier) of the POLYCOST database, we collected the speech of 15 male speakers and 15 female speakers. We pooled the feature vectors from all these 30 speakers and prepared three types of UBM (i.e., UBM of 512 mixtures, UBM of 256 mixtures and UBM of 128 mixtures) by using the iterative expectation maximization (EM) algorithm [Bilmes, J.A., 1998]. We have used the remaining 100 speakers for training and testing. To build the final speaker model, for each speaker, we have used two minutes of training data (i.e., MFCC feature vectors extracted from the two minutes of speech) to adapt the UBM. For MAP adaptation of UBM, we chose the relevance factor, $r = 16$. At the time of MAP adaptation, only the mean vectors of the UBM have been adapted [Reynolds, D.A., Quatieri, T.F., & Dunn, R.B., 2000]. The test data has been tested with all speakers. Hence, for each test utterance, there is one true score and ninety-nine false scores. At the time of testing, we have applied the test segments of durations 20 seconds, 10 seconds and 5 seconds.

### 3.3 Preparation of GMM-SVM Classifier

We have trained the GMM-SVM classifier with GMM supervector as the linear kernel. Initially, we have not performed partitioning of training utterance (i.e., Section 4). Therefore, a complete training speech of 120 seconds was used to adapt the UBM. Later, we have performed partitioning of training utterance and adapted the UBM with each partition separately (i.e., Section 5). We have chosen 19 dimensional MFCC feature vector. Therefore, for UBM of 512 mixtures, the size of each GMM supervector becomes 9728 (i.e., 19 multiplied by 512). Similarly, for UBM of 256 mixtures, the size of each GMM supervector becomes 4864. In the same way, for UBM of 128 mixtures, the size of each GMM supervector becomes 2432. We have trained the SVM classifier with soft-margin decision boundary by using one against the rest approach [Burges, C. J. C., 1998]. For preparation of SVM classifier, we have used the LIBSVM software [Chang, C.-C., Lin, C.-J., 2001].

# 4 COMPARISON OF GMM-UBM CLASSIFIER WITH GMM-SVM CLASSIFIER FOR RECOGNITION OF LONG DURATION AND SHORT DURATION TEST SEGMENTS

In this section, we are comparing the recognition performance of GMM-UBM classifier with GMM-SVM classifier. Here, for the training of GMM-SVM classifier we have not applied partitioning of training utterance approach. Table 1 shows the equal error rate (EER %) performances of GMM-UBM classifiers and GMM-SVM classifiers for different model orders (i.e., 512, 256 & 128) and supervector dimensions (i.e., 9728, 4864 & 2432). For test speech segments of duration 20 seconds the number of true trials is 1422 and the number of impostor trials is 140778. Similarly, for test speech segments of duration 10 seconds the number of true trials is 2888 and the number of impostor trials is 285912. Lastly, for test speech segments of duration 5 seconds the number of true trials is 5827 and the number of impostor trials is 576873.

Table 1 Comparison of equal error rate (EER %) performances between GMM-UBM classifier and GMM-SVM classifier for three different UBM model orders and three different durations of test speech segments.
MO= Model order of UBM, SD= Supervector dimension

|  | Duration of Test Speech Segments | | |
|---|---|---|---|
|  | 20 s | 10 s | 5 s |
| GMM-UBM (MO=512) | 7.19 % | 7.38 % | 7.97 % |
| GMM-SVM (SD=9728) | 5.98 % | 9.18 % | 17.90 % |
| GMM-UBM (MO=256) | 7.17 % | 7.55 % | 7.96 % |
| GMM-SVM (SD=4864) | 5.98 % | 8.28 % | 15.86 % |
| GMM-UBM (MO=128) | 7.03 % | 7.56 % | 8.01 % |
| GMM-SVM (SD=2432) | 6.54 % | 7.86 % | 13.15 % |

Fig. 2 shows comparison of performances between GMM-UBM classifier of model order 512 and GMM-SVM classifier with GMM supervector of dimension 9728 for recognition of test speech segments of long duration and short duration. From Fig. 2 it is clear that, for test speech segments of long duration (i.e., 20 s), irrespective of the operating points the GMM-SVM classifier performed significantly better than GMM-UBM classifier. However, for test speech segments of short duration (i.e., 5 s), the performance of GMM-SVM classifier is very poor. From Fig. 2 it is clear that, for test speech segments of short duration (i.e., 5 s), irrespective of the operating points the GMM-UBM classifier performed far better than the GMM-SVM classifier.

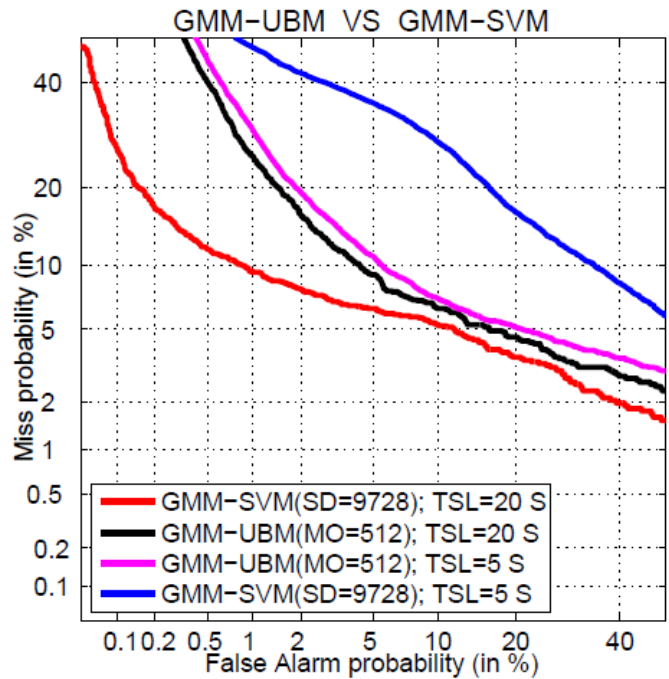

Fig. 2 DET curves showing the comparison of speaker recognition performances between GMM-UBM classifier of model order 512 and GMM-SVM classifier with GMM supervector of dimension 9728 for recognition of test speech segments of long and short durations.

Fig. 3 shows the effect of dimension of the GMM supervector (SD) on the performance of GMM-SVM classifier for recognition of short duration test speech segments (i.e., 5 s). From Fig. 3 it is clear that, irrespective of the operating points, for recognition of short duration test speech segments the performance of GMM-SVM classifier decreases as the dimension of the GMM supervector increases. However, dimension of the MFCC feature vector is fixed (here it is 19). Hence, dimension of the GMM supervector is higher which implies that, the model order of the UBM is higher. Therefore, it is very important to select the proper model order of the UBM to generate the GMM supervector.

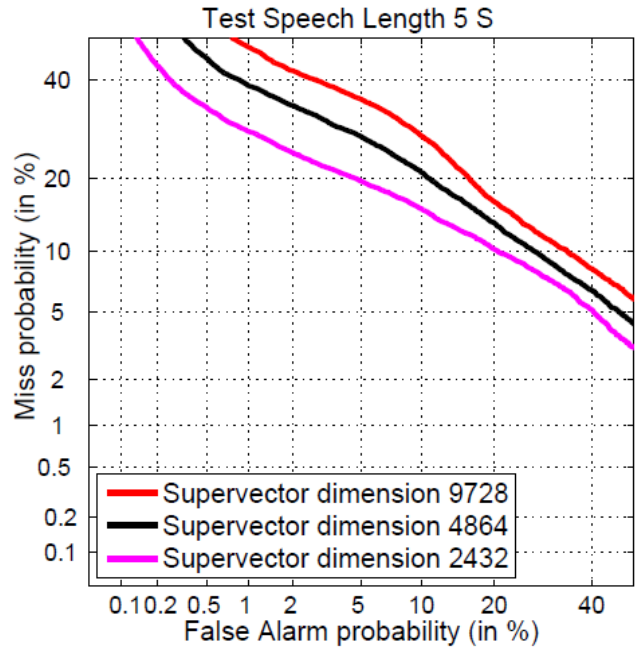

Fig. 3 DET curves showing the effect of the dimension of GMM supervector on the performance of GMM-SVM classifier for recognition of test speech segments of short duration.

From the above results (Table 1, Fig. 2 and Fig. 3) we draw the following five conclusions:

(1) *When the training speech is sufficient, the degradation in recognition performance due to the reduction of duration of test segment (i.e., from 20 seconds to 5 seconds) is negligible in classical GMM-UBM classifier compared to the GMM-SVM classifier (see* Fig. 2*).*

(2) *For long duration test segments (i.e., 20 s), the recognition performance of GMM-SVM classifier is better than GMM-UBM classifier (see* Fig. 2*). Hence, for test segments of long duration (i.e., 20 s), our experimental results give similar observations as found by other researchers on different data bases [Dehak & Chollet, 2006; Campbell,W.M., Sturim, D.E., & Reynolds, D.A., 2006; Campbell,W.M., Sturim, D.E., Reynolds, D.A., & Solomonoff, A., 2006].*

(3) *For short duration test segments (i.e.,5 s), the recognition performance of GMM-SVM classifier is very poor compared to the classical GMM-UBM classifier (see* Fig. 2*).*

*Therefore, for test segments of short duration (i.e., 5 s), it is not feasible to use GMM-SVM classifier in a usual configuration.*

(4) *Data imbalance problem is not responsible for the poor performance of GMM-SVM classifier for recognition of test segments of short duration.*

(5) *If dimension of GMM supervector increases, then performance of GMM-SVM classifier for recognition of short duration test segments degrades further (see Fig. 3).*

The conclusion (1) essentially says the following:

*When we have adequate amount of training data, the degradation in recognition performance due to the reduction in duration of test segment is small in GMM-UBM classifier. The Intuitive explanation for this conclusion is given below:*

Training feature vectors and test feature vectors of the same speaker stay closely in the feature space (we perform speaker recognition task based on this property). For same phone (a phone is the distinct sound unit, it is the physical existence of phoneme. We can also say that a phoneme is an underlying object whose surface representations are phones.), training and test feature vectors of the same speaker stay very close together in the feature space. However, for different phones, training and test feature vectors of the same speaker stay a little apart in the feature space. In other words, feature vectors generated from a speaker form various clusters depending on the emitted phones. We call these clusters the "acoustic zones". We first take a detailed relook at the calculation of average log-likelihood ratio score for GMM-UBM classifier for the test segments as follows:

The sequence $\mathbf{Y} = \{\mathbf{y}_1, \mathbf{y}_2, \cdots, \mathbf{y}_T\}$ represents the total feature vectors extracted from the test segments. The set $\mathbf{Y}$ can be segregated into two sets as follows,

$$\mathbf{Y} = \mathbf{Y}_S \cup \mathbf{Y}_{NS}, \qquad (2)$$

Where, set $\mathbf{Y}_S$ contains the test feature vectors which belong to the same "acoustic zones" as occupied by the training feature vectors of that speaker (i.e., types of phones contained in these portions of test speech are also present in the training speech). Set $\mathbf{Y}_{NS}$ contains the test feature vectors which do not belong to the "acoustic zones" of the training feature vectors (i.e., types of phones contained in these portions of test speech are not present in the training speech). To simplify calculation, it is generally assumed that, the feature vectors of set $\mathbf{Y}$ are independent. Therefore, the average log-likelihood score calculation will be as follows:

$$\Lambda(\mathbf{Y}) = \frac{1}{T}\log(\frac{p(\mathbf{Y}|\lambda_{spk})}{p(\mathbf{Y}|\lambda_{UBM})}) = \frac{1}{T}\log(\frac{p(\mathbf{y}_1|\lambda_{spk})}{p(\mathbf{y}_1|\lambda_{UBM})}\frac{p(\mathbf{y}_2|\lambda_{spk})}{p(\mathbf{y}_2|\lambda_{UBM})}\cdots\frac{p(\mathbf{y}_T|\lambda_{spk})}{p(\mathbf{y}_T|\lambda_{UBM})}). \qquad (3)$$

In the adaptation of the speaker model $\lambda_{spk}$ from the universal background model $\lambda_{UBM}$, mixture parameters for those "acoustic zones" which are not seen in the training speech are simply copied from the $\lambda_{UBM}$. Hence, the test feature vectors which do not belong to the "acoustic zones" occupied by the training feature vectors will produce approximately same likelihood values for speaker model $\lambda_{spk}$ and background model $\lambda_{UBM}$. As a result, during recognition, the feature vectors belonging to the "acoustic zones" unseen in the speaker's training speech do not contribute in the average log-likelihood ratio score. In other words, types of phones contained in the test speech which are not present in the training speech of that speaker, do not contribute in the average log-likelihood ratio score calculation. Therefore, we have the following two important conclusions:

$$\Lambda(\mathbf{y}_t) = \log(\frac{p(\mathbf{y}_t|\lambda_{spk})}{p(\mathbf{y}_t|\lambda_{UBM})}) \cong 0 \qquad \text{for} \quad \mathbf{y}_t \in \mathbf{Y}_{NS} \qquad (4)$$

$$\Lambda(\mathbf{Y}) = \frac{1}{T}\log(\frac{p(\mathbf{Y}|\lambda_{spk})}{p(\mathbf{Y}|\lambda_{UBM})}) = \frac{1}{T}\sum_{\mathbf{y}_t \in \mathbf{Y}_S}\log(\frac{p(\mathbf{y}_t|\lambda_{spk})}{p(\mathbf{y}_t|\lambda_{UBM})}) \qquad (5)$$

We define; the speaker model $\lambda_{spk}$ is well trained if following condition holds:

$$\tau = \frac{\text{Total number of test feature vectors belonging to set } \mathbf{Y}_{NS}}{\text{Total number of test feature vectors belonging to set } \mathbf{Y}_{S}} << 1 \qquad (6)$$

If the amount of training data is large, then this represents the situation where evidences of various "acoustic zones" collected from the training speech of the speaker are sufficient (i.e., training data contains different types of phones). Therefore, the adapted speaker model $\lambda_{spk}$ will represent the corresponding speaker quite well and this implies that the speaker model $\lambda_{spk}$ is well trained. At the time of training of speaker model, we have used two minutes of training speech (the two minutes of speech are taken after removal of silence portions). It is generally assumed that, this duration of speech data is adequate to represent the speaker [Reynolds, D.A., Quatieri, T.F., and Dunn, R.B., 2000]. Therefore, we conclude that, the speaker model $\lambda_{spk}$ is well trained.

At the time of testing, test segment of long duration (20 s) represents 2000 feature vectors and test segment of short duration (5 s) represents 500 feature vectors. We assume that 10 % of total test feature vectors belong to the set $\mathbf{Y}_{NS}$. Then the average log-likelihood ratio score calculation for long duration and short duration test segments will be as follows:

$$\Lambda_{20}(\mathbf{Y}) = \frac{1}{2000} \sum_{\substack{t=1 \\ \mathbf{y}_t \in \mathbf{Y}_S}}^{1800} \log(\frac{p(\mathbf{y}_t | \lambda_{spk})}{p(\mathbf{y}_t | \lambda_{UBM})}) \quad \text{and} \quad \Lambda_{5}(\mathbf{Y}) = \frac{1}{500} \sum_{\substack{t=1 \\ \mathbf{y}_t \in \mathbf{Y}_S}}^{450} \log(\frac{p(\mathbf{y}_t | \lambda_{spk})}{p(\mathbf{y}_t | \lambda_{UBM})}), \qquad (7)$$

Where $\Lambda_{20}(\mathbf{Y})$ is the mean score for test segments of long duration (20 s) and $\Lambda_{5}(\mathbf{Y})$ is the mean score for test segments of short duration (5 s). When, the model is well trained then, the term $\log p(\mathbf{y}_t | \lambda_{spk}) - \log p(\mathbf{y}_t | \lambda_{UBM})$ always gives strong evidence toward the hypothesized speaker for $\mathbf{y}_t \in \mathbf{Y}_S$. As a result, the operation mean will normalize the effect of different durations of the test speech segments. Therefore, the two mean values (i.e., $\Lambda_{20}(\mathbf{Y})$ and $\Lambda_{5}(\mathbf{Y})$) will give approximately the same accuracy.

However, if the amount of training data is not adequate then the speaker model $\lambda_{spk}$ will not be well trained. As a result, lots of test feature vectors will arise from various "acoustic zones" which are not seen in the training speech (i.e., $\mathbf{y}_t \in \mathbf{Y}_{NS}$ will be large). Hence, the mean log-likelihood ratio score calculation for test segments of short duration is adversely affected. Therefore, the accuracy of recognition for test segments of short duration will be very poor. In this case (i.e., when the speaker model is not well trained) the performance of recognition for test segments of long duration will also degrade. However, the degradation of accuracy for test segments of long duration is relatively less compared to the degradation of accuracy for test segments of short duration. Because it is highly probable that, for test segments of long duration, some test feature vectors arise from "acoustic zones" which are occupied by the training feature vectors of that speaker (i.e., types of phones present in some portions of test speech are also present in the training speech). On the other hand, for test segments of short duration, frequently, none of the test feature vectors arise from "acoustic zones" which are occupied by the training feature vectors of that speaker (i.e., types of phones present in test speech are recurrently not present in the training speech). In fact, the works carried out by Fauve et al. and Kinnunen et al. have shown that, the accuracy of GMM-UBM classifier reduces drastically for short duration training and short duration testing case compared to long duration training and long duration testing case [Fauve, B., Evans, N., Pearson, N., Bonastre, J.-F., & Mason, J., 2007; Kinnunen, T., Saastamoinen, J., Hautamaki, V., Vinni, M., & Franti, P., 2009].

The conclusions (2) & (3) essentially mean the following:

> *For sufficient training speech, the degradation in performance due to the reduction in the duration of test segment is very large in GMM-SVM classifier. The Intuitive explanation for this conclusion is given below:*

It is known that; the speaker model is generated by adapting the mean vectors of the UBM by using the MFCC feature vectors extracted from the speaker's speech. For easy understanding, we write the following relations which are used at the time of MAP adaptation to generate the speaker model from the UBM. These relations are well explained in the literature [Reynolds, D.A., Quatieri, T.F., & Dunn, R.B., 2000].

$$n_i = \sum_{t=1}^{T} \Pr(i|\mathbf{x}_t), \qquad \alpha_i = \frac{n_i}{n_i + r}, \qquad \boldsymbol{\mu}_i = \alpha_i E_i(\mathbf{x}) + (1-\alpha_i)\boldsymbol{\mu}_i \qquad (8)$$

From the above equation, it is clear that, if the number of feature vectors, near the $i^{th}$ Gaussian is large, then the sufficient statistics of weight for the $i^{th}$ Gaussian $n_i$ is also large. As a result, the data-dependent adaptation coefficient for the $i^{th}$ Gaussian $\alpha_i$ is close to unity (i.e., $\alpha_i \rightarrow 1$) and the adapted mean vector for the $i^{th}$ Gaussian $\boldsymbol{\mu}_i$ depends mainly on the new sufficient statistics of mean vector $E_i(\mathbf{x})$ derived from the feature vectors of the speaker's speech. However, if the number of feature vectors, near the $i^{th}$ Gaussian is less, then the sufficient statistics of weight for the $i^{th}$ Gaussian $n_i$ is also small. Thus, the data-dependent adaptation coefficient for the $i^{th}$ Gaussian is much less than one (i.e., $\alpha_i << 1$) and the adapted mean vector for the $i^{th}$ Gaussian $\boldsymbol{\mu}_i$ depends mainly on the value of the mean vector of the UBM, $\boldsymbol{\mu}_i$.

At the time of generation of the training supervector, we have used two minutes of training utterance (this duration of speech is calculated after removal of silence portions). It implies that, total 12000 MFCC feature vectors were used to generate the training GMM supervector (frame size of 20 milliseconds with 50 % overlap). As a result, some mean vectors of the UBM which were nearer to the "acoustic zone" of the speaker's speech changed significantly to generate the training GMM supervector. However, at the time of testing if the duration of test speech is small, then the test GMM supervector is created by adapting the UBM with small amount of test feature vectors. For example, test segment of duration 5 seconds implies that, total 500 MFCC feature vectors were used to generate the test GMM supervector. As a result, those mean vectors of the UBM which were nearer to the acoustic space of the speaker's speech changed very less (i.e., compared to the training condition) to generate the test GMM supervector. Therefore, there is a large difference between training GMM supervector and test GMM supervector for the same speaker. We postulate that, this large mismatch of the amount of MAP adaptations between training GMM supervector and test GMM supervector is the inherent cause for poor performance of GMM-SVM classifier for recognition of test segments of short duration. Pictorial depictions of our postulation are given below:

Fig. 4 shows the preparation of training GMM supervector of a hypothesized speaker with the help of MAP adaptation from UBM. Here, two Gaussians of the UBM are nearer to the training data. Therefore, the mean vectors of those two Gaussians have changed significantly. However, the mean vectors of remaining two Gaussians have not changed because they are far away from the training data. From the mean vectors of the Fig. 4 (b) we generate the training GMM supervector.

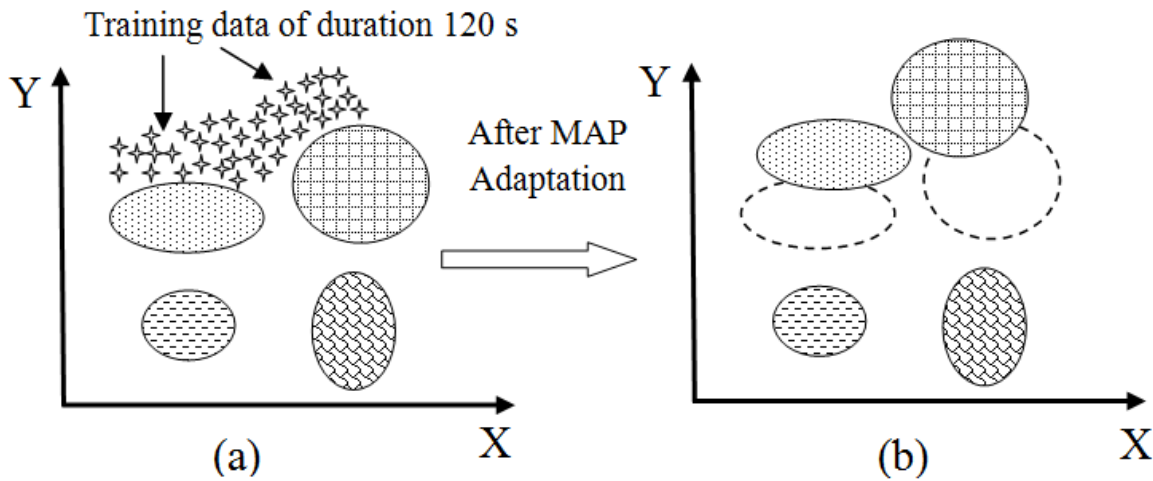

Fig. 4 Preparation of the training GMM supervector using MAP adaptation from UBM. Here, 4 (a) indicates the condition before MAP adaptation and 4 (b) indicates the condition after MAP adaptation.

Fig. 5 shows the preparation of test GMM supervector for that hypothesized speaker for test segments of duration 20 seconds. Here also, two Gaussians of the UBM are nearer to the test data. However, the amount of test data is much less compared to the training case of Fig. 4. Therefore, the amounts of shifts of the mean vectors of those two Gaussians of the UBM are also much less compared to the training case. From the mean vectors of the Fig. 5 (b) we generate the test GMM supervector of duration 20 seconds.

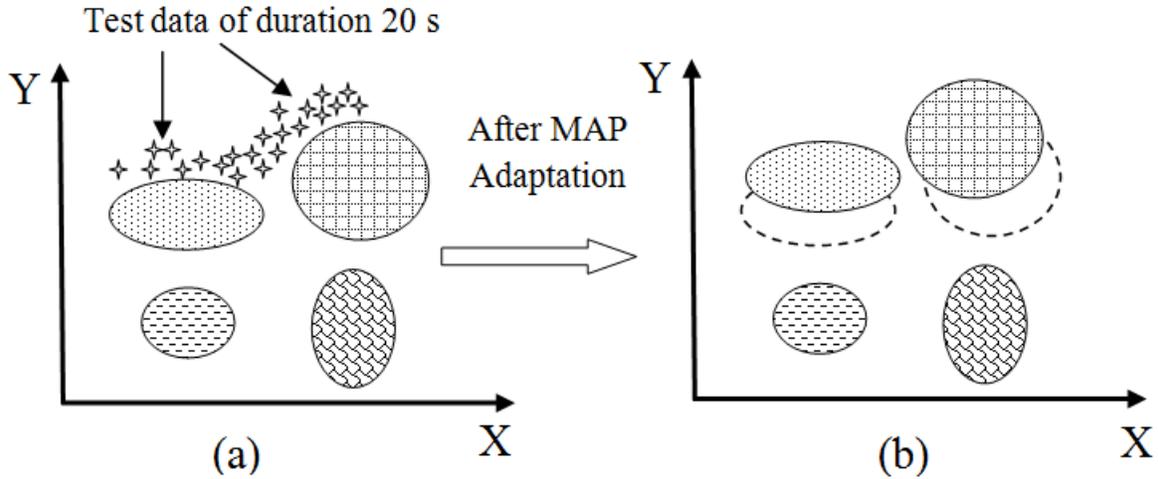

Fig. 5 Preparation of the test GMM supervector using MAP adaptation from UBM for test speech of duration 20 seconds. Here, 5 (a) indicates the condition before MAP adaptation and 5 (b) indicates the condition after MAP adaptation.

Fig. 6 shows the preparation of test GMM supervector for the same hypothesized speaker for test segments of duration 5 seconds. Here, the amount of test data is very less compared to the training case of Fig. 4. Therefore, the amounts of shifts of the mean vectors of those two Gaussians of the UBM are also very less compared to the training case. From the mean vectors of the Fig. 6 (b) we generate the test GMM supervector of duration 5 seconds.

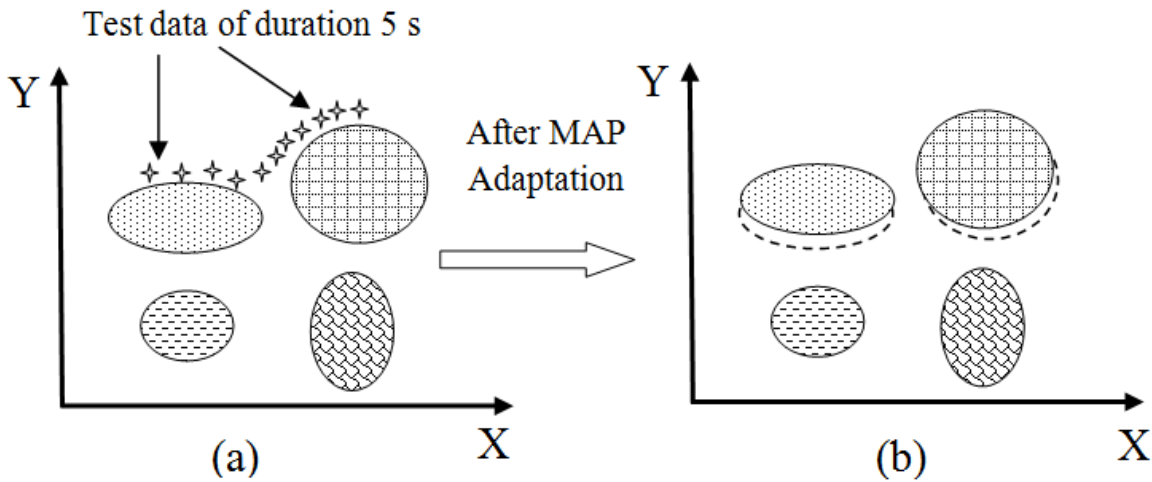

Fig. 6 Preparation of the test GMM supervector using MAP adaptation from UBM for test speech of duration 5 seconds. Here, 6 (a) indicates the condition before MAP adaptation and 6 (b) indicates the condition after MAP adaptation.

Now it is clear from Fig.4 (b) and Fig.6 (b) that, there is a large difference between the amounts of MAP adaptations of training GMM supervector and test GMM supervector for the same hypothesized speaker when the duration of test speech is very less compared to the duration of training speech. Therefore, at the time of testing, the false rejection rate is very high. Due to this high false rejection rate, GMM-SVM classifier performs very poorly for recognition of short duration test segments.

*The conclusion (4) rejects the possibility of data imbalance problem as the cause for the poor performance of GMM-SVM classifier for the recognition of test segments of short duration. Therefore, conclusion (4) essentially rejects the hypothesis which was proposed by Mak, M.W., & Rao,W. in 2011 to explain the reason for the poor performance of GMM-SVM classifier. The intuitive explanation for conclusion (4) is given below:*

Data imbalance problem refers to the phenomenon when the training samples of positive class are very less compared to the training samples of the negative class. As a result, the decision boundary shifts towards the positive class. Therefore, the data imbalance problem and its associated effect (i.e., shifting of decision boundary towards the positive class) are completely related to the training condition of the classifier. If the data imbalance problem occurs at the time of training then at the time of testing, the number of false negatives may increase due to the shifting of decision boundary towards the positive class. As a result, the accuracy of the discriminative classifier may degrade.

It is clear from Table 1 and Fig. 2 that, the GMM-SVM classifier performs quite well for recognition of test segments of long duration and very poorly for recognition of test segments of short duration. If the data imbalance problem is the cause for the poor performance of the GMM-SVM classifier for the recognition of test segments of short duration then it should also affect the recognition performance of the GMM-SVM classifier for test segments of long duration, because, data imbalance problem is completely related with the training condition of the classifier. However, it is clear that, the recognition performance of the GMM-SVM classifier has not been adversely affected for test segments of long duration. Therefore, we conclude that, the data imbalance problem is not responsible for the poor performance of the GMM-SVM classifier for the recognition of test segments of short duration.

*The conclusion (5) reveals the relation between the dimension of the GMM supervector and the recognition performance of the GMM-SVM classifier. Conclusion (5) essentially says that, we should not choose the dimension of the GMM supervector unnecessarily higher. We will give the detailed intuitive explanation for this conclusion in section* 6 *and in section* 7.

# 5 EFFECT OF PARTITIONING OF TRAINING UTTERANCE ON GMM-SVM CLASSIFIER FOR RECOGNITION OF TEST SEGMENTS OF LONG DURATION AND TEST SEGMENTS OF SHORT DURATION

In this section, we investigate the effects of partitioning of training utterance (discussed in Section 2) on the performance of GMM-SVM classifier for recognition of test segments of long duration (i.e., 20 s) and test segments of short duration (i.e., 5 s). Table 2, Table 3 and Table 4 show the effect of partitioning of training utterance (UP) on the equal error rate (EER %) performances of GMM-SVM classifiers with GMM supervectors of dimension 9728, 4864 and 2432 respectively.

Table 2 Effect of partitioning of training utterance (UP) on equal error rate (EER %) performances of GMM-SVM classifier with GMM supervector of dimension 9728 for recognition of test speech segments of different durations.

| No of Partitions of Training Utterance | Duration of Test Speech Segments | | |
|---|---|---|---|
| | 20 s | 10 s | 5 s |
| NO | 5.98 % | 9.18 % | 17.90 % |
| 2 | 5.27 % | 6.30 % | 11.31 % |
| 3 | 5.34 % | 5.78 % | 8.57 % |
| 4 | 5.41 % | 5.75 % | 7.29 % |
| 6 | 5.42 % | 5.40 % | 6.21 % |
| 8 | 5.34 % | 5.30 % | 5.83 % |
| 12 | 5.63 % | 5.57 % | 5.87 % |

Table 3 Effect of partitioning of training utterance (UP) on equal error rate (EER %) performances of GMM-SVM classifier with GMM supervector of dimension 4864 for recognition of test speech segments of different durations.

| No of Partitions of Training Utterance | Duration of Test Speech Segments | | |
|---|---|---|---|
| | 20 s | 10 s | 5 s |
| NO | 5.98 % | 8.28 % | 15.86 % |
| 2 | 5.40 % | 6.03 % | 10.45 % |
| 3 | 5.27 % | 5.71 % | 8.20 % |
| 4 | 5.41 % | 5.68 % | 7.06 % |
| 6 | 5.20 % | 5.51 % | 6.26 % |
| 8 | 5.20 % | 5.40 % | 6.08 % |
| 12 | 5.35 % | 5.37 % | 5.78 % |

Table 4 Effect of partitioning of training utterance (UP) on equal error rate (EER %) performances of GMM-SVM classifier with GMM supervector of dimension 2432 for recognition of test speech segments of different durations.

| No of Partitions of Training Utterance | Duration of Test Speech Segments | | |
|---|---|---|---|
| | 20 s | 10 s | 5 s |
| NO | 6.54 % | 7.86 % | 13.15 % |
| 2 | 5.84 % | 6.51 % | 9.23 % |
| 3 | 5.91 % | 6.37 % | 8.08 % |
| 4 | 5.70 % | 6.13 % | 7.29 % |
| 6 | 5.77 % | 5.99 % | 6.95 % |
| 8 | 5.41 % | 5.92 % | 6.71 % |
| 12 | 5.34 % | 5.64 % | 6.69 % |

From the results of Table 2 to Table 4 and from Fig. 7 it is clear that, the recognition performances of the GMM-SVM classifiers for test segments of long duration (i.e., 20 s) have improved marginally after partitioning of the training utterance appropriately. It is also evident from Fig. 7 that; noticeable improvement occurs when we partition the training utterance by two. After that, the overall performance of the GMM-SVM classifier (i.e., if we consider all the operating points of the DET curves) almost saturates. In our experimental setup, training utterance partitioned by two implies that, the length of each sub utterance used for training is 60 seconds. This observation indicates that, the length of test segments should not be less than one third compared to the length of the training utterance.

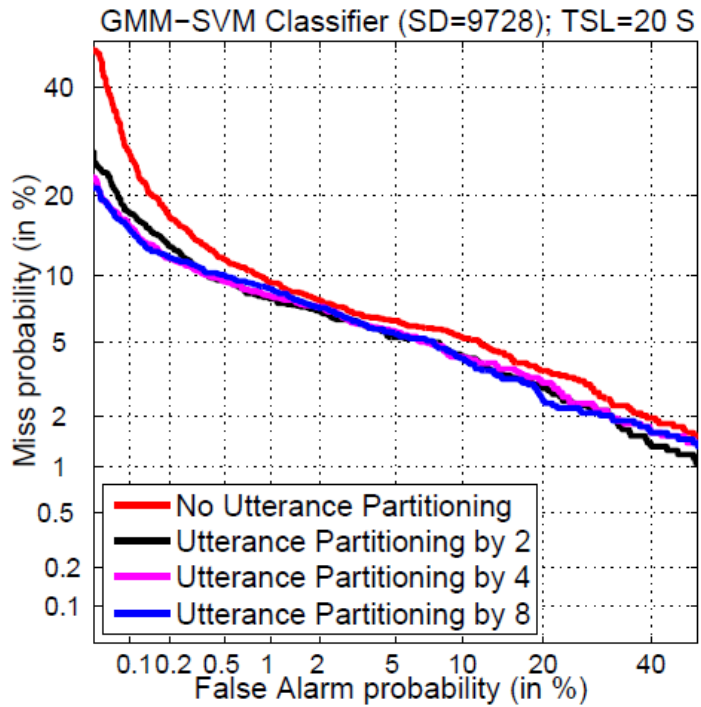

Fig.7 Effect of partitioning of training utterance (UP) on the performance of GMM-SVM classifier with GMM supervector of dimension (SD) 9728 for recognition of test segments of long duration (20 s). Here, TSL=test segment length.

From the results of Table 2 to Table 4 and from Fig. 8 and Fig. 9 it is clear that, the recognition performances of the GMM-SVM classifiers for test segments of short duration (i.e., 5 s) have improved very significantly after partitioning of the training utterance appropriately.

Initially the overall performance of the GMM-SVM classifier increases rapidly due to the partitioning operation. However, the performance of the GMM-SVM classifier saturates when we partition the training utterance by eight. In our experimental setup, training utterance partitioned by eight implies that, the length of each sub utterance used for training is 15 s. Therefore, for short duration test segments also, we observe that, the length of test segments should not be less than one third compared to the length of training utterance.

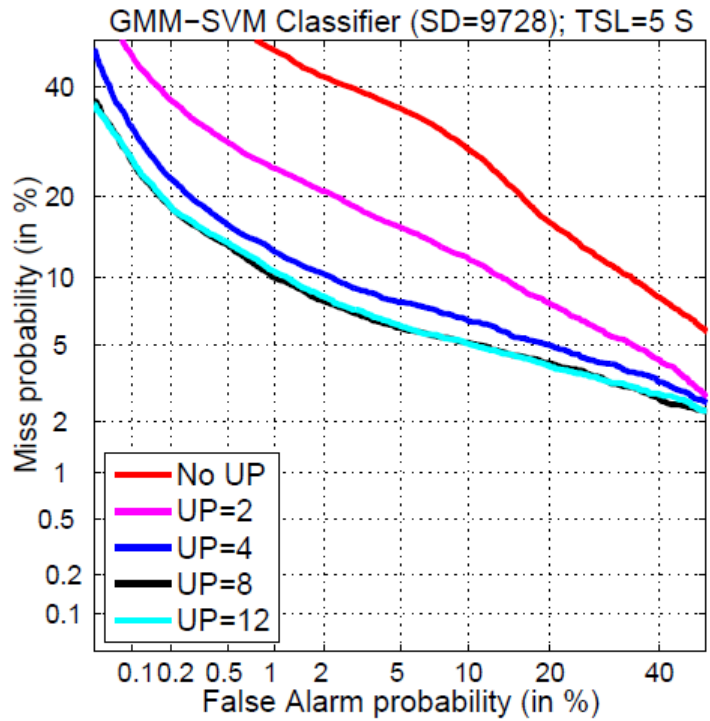

Fig. 8 Effect of partitioning of training utterance (UP) on the performance of GMM-SVM classifier with GMM supervector of dimension (SD) 9728 for recognition of test segments of short duration (5 s). Here, TSL=test segment length.

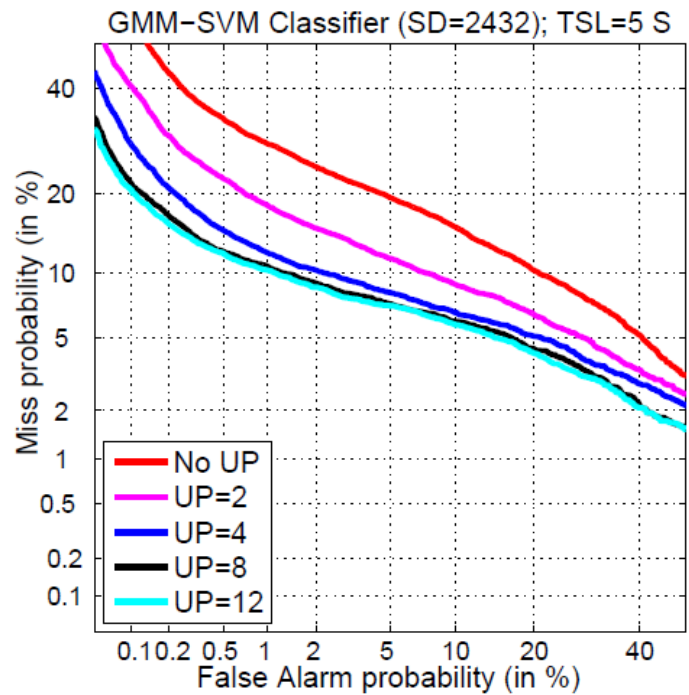

Fig. 9 Effect of partitioning of training utterance (UP) on the performance of GMM-SVM classifier with GMM supervector of dimension (SD) 2432 for recognition of test segments of short duration (5 s). Here, TSL=test segment length.

Based on the above results (i.e., Table 1 to Table 4) it is clear that, after partitioning of training utterances GMM-SVM classifier performs much better than GMM-UBM classifier for recognition of test segments of short duration also. DET curves in Fig. 10 give the comparison between GMM-UBM classifier of model order 512 with GMM-SVM classifier with GMM supervector of dimension 9728 after training utterances have been partitioned by eight for recognition of test segments of short duration (i.e., 5 s). It is evident from Fig. 10 that, after we partition the training utterances by eight, the GMM-SVM classifier performed significantly better compared to the GMM-UBM classifier irrespective of the operating points.

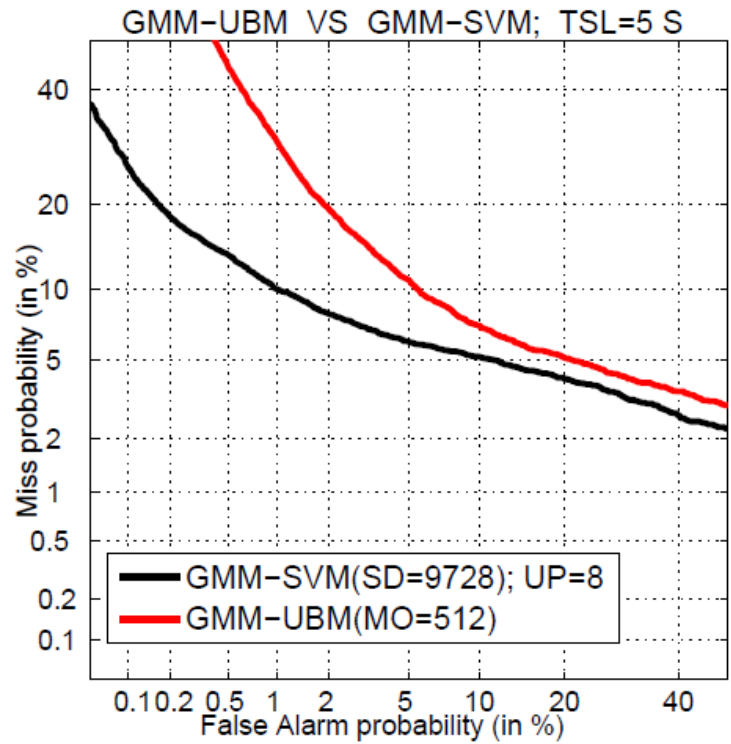

Fig. 10 DET curves showing the comparison between GMM-UBM classifier of model order (MO) 512 and GMM-SVM classifier with GMM supervector of dimension (SD) 9728 after training utterances have been partitioned by eight for recognition of test segments of short duration (5 s). Here, TSL=test segment length.

Table 5 shows the duration of each sub-utterance at the time of application of training utterance partitioning approach. It also shows the number of GMM supervectors in the positive class and number of GMM supervectors in the negative class, when we apply the training utterance partitioning (UP) approach for training of the GMM-SVM classifier. It is obvious from Table 5 that, after the application of training utterance partitioning (UP) approach, the ratio of the data imbalance between positive class and negative class (i.e., PCS:NCS) has not changed. Therefore, the data imbalance problem of the GMM-SVM classifier has not reduced after the application of training utterance partitioning approach.

Table 5 Duration of each sub-utterance and number of GMM supervectors in the positive class and number of GMM supervectors in the negative class after application of the training utterance partitioning approach.
PCS=Positive Class Supervectors. NCS=Negative Class Supervectors.

| No of Partitions of the Training Utterance | Duration of Each Sub-Utterance | Number of GMM Supervectors in the Positive Class (PCS) | Number of GMM Supervectors in the Negative Class (NCS) | Ratio of the Data Imbalance (PCS: NCS) |
|---|---|---|---|---|
| NO | 120 s | 1 | 99 | 1:99 |
| 2 | 60 s | 2 | 198 | 1:99 |
| 3 | 40 s | 3 | 297 | 1:99 |
| 4 | 30 s | 4 | 396 | 1:99 |
| 6 | 20 s | 6 | 594 | 1:99 |
| 8 | 15 s | 8 | 792 | 1:99 |
| 12 | 10 s | 12 | 1188 | 1:99 |

From the above results (Table 1 to Table 5 and Fig. 7 to Fig. 10) and discussions, we derive the following six conclusions:

(6) *Application of the training utterance partitioning approach (UP) is not reducing the data imbalance problem of the GMM-SVM classifier.*

(7) *Performance of GMM-SVM classifier for recognition of test segments of long duration (i.e., 20 s) has improved marginally after the application of partitioning of training utterance (UP) approach.*

(8) *Performance of GMM-SVM classifier for recognition of test segments of short duration (i.e., 5 s) has improved very significantly after the application of partitioning of training utterance (UP) approach.*

(9) *Conclusions (6), (7) and (8) corroborate to our previous conclusion number (4) which tells that, data imbalance problem is not responsible for the poor performance of the GMM-SVM classifier.*

(10) *When we partition the training utterance by a suitable number, the GMM-SVM classifier performs far better than the classical GMM-UBM classifier.*

(11) *The duration of the test speech segments should not be less than one third of the duration of the training speech utterance.*

The intuitive explanation of conclusions (6) is given below:

When we apply partitioning of the training utterance (UP) approach for training of the GMM-SVM classifier then, positive class gets more number of supervectors compare to the situation if we do not apply partitioning of the training utterance approach for training. However, the negative class also gets more number of supervectors, because, we are applying UP approach for positive class as well as for negative class (see Fig. 1). Therefore, the ratio between the numbers of supervectors of positive class to the numbers of supervectors of negative class (i.e., PCS:NCS) is same before the application of UP approach and after the application of UP approach for training of the GMM-SVM classifier (see Table 5). Therefore, partitioning of the training utterance approach is not reducing the data imbalance problem of the GMM-SVM classifier (see Table 5). Hence, we strongly reject the hypothesis which was proposed by Mak, M.W., & Rao,W., in 2011, because, according to their proposition, after application of the training utterance partitioning approach, the performance of the GMM-SVM classifier improved due to the reduction of the data imbalance problem.

The intuitive explanation of conclusions (7) and (8) is given below:

We have observed in section 4 that (see Table 1 and Fig. 2), the performance of the GMM-SVM classifier for recognition of test segments of long duration (i.e., 20 s) is quite good. However, the performance of the GMM-SVM classifier for recognition of test segments of short duration (i.e., 5 s) is very poor. We have explained these observations by the phenomenon of the mismatch between the amounts of MAP adaptations for training GMM supervector and test GMM supervectors for the same speaker (see section 4). In this experimental review and analysis article we have postulated that (see Fig. 4 to Fig. 6), when the duration of test segment is much less than the duration of training utterance then, there is a large mismatch of the amounts of MAP adaptations between training GMM supervector and test GMM supervector for the same speaker. Hence, there are many false negatives at the time of testing. Numbers of false negatives increase if the duration of test speech decreases than the duration of the training utterance (because mismatch of the amounts of MAP adaptations increases). Therefore, numbers of false negatives at the time of testing are much less for test segments of long duration (i.e., 20 s); however, false negatives are very significantly prominent at the time of testing for test segments of short duration (i.e., 5 s).

When we apply partitioning of the training utterance (UP) approach for training of the GMM-SVM classifier, then we use each partition to generate a separate GMM supervector. We use all these supervectors at the time of training of the classifier (see section 2). Each partition contains much less numbers of training feature vectors compared to the total numbers of the training feature vectors generated from the complete training utterance. Therefore, the mismatch of the amounts of MAP adaptations between training GMM supervector and test GMM supervector for the same speaker is much less when we apply partitioning of the training utterance approach for creation of the training GMM supervectors. Hence, Numbers of false negatives are also much less when we apply partitioning of the training utterance approach for training of the GMM-SVM classifier. However, when we compare Fig. 4 and Fig. 5 with Fig.4 and Fig. 6, it is clear that, the scope for improvement of the recognition accuracy of the GMM-SVM classifier by reducing the numbers of false negatives after the application of UP approach is much more for test segments of short duration (i.e., 5 s) compared to the test segments of long duration (i.e., 20 s), because, mismatch of the amounts of MAP adaptations between training GMM supervector and test GMM supervectors for the same speaker is also much more for test segments of short duration compared to the test segments of long duration. Hence, after the application of partitioning of training utterance (UP) approach, the performance of GMM-SVM classifier for recognition of test segments of short duration (i.e., 5 s) has improved very significantly, because, the scope for reduction of false negatives by application of UP approach is very high for test segments of short duration. However, the performance of GMM-SVM classifier for recognition of test segments of long duration (i.e., 20 s) has improved marginally, because, the scope for reduction of false negatives by application of UP approach is much less for test segments of long duration. Therefore, conclusion (7) and conclusion (8) are very logical.

The intuitive explanation of conclusion (9) is given below:

In their work, Mak & Rao [Mak, M.W., & Rao,W., 2011] did not show the effects of the duration of the test segments on the recognition performance of the GMM-SVM classifier. In this work, we have explored the effects of the duration of the test segments on the recognition performance of the GMM-SVM classifier before the application of UP approach as well as after the application of UP approach. In this experimental review and analysis article, we have clearly shown that, data imbalance problem of the GMM-SVM classifier has not reduced after the application of the training utterance partitioning approach (see Table 5). Hence, the recognition performances of the GMM-SVM classifier have improved after the application of training utterance partitioning (UP) approach without reducing the data imbalance problem. Therefore, it confirms that, when we do not apply the UP approach, the reason for poor recognition performance of the GMM-SVM classifier for test segments of short duration is not the data imbalance problem of the GMM-SVM classifier. Mak & Rao postulated that, more numbers of supervectors present in the positive class after the application of UP approach shift the decision hyperplane towards the negative class, hence, recognition performance increases. However, they neglected the facts that, at the time of application of UP approach, the number of GMM supervectors of the negative class also increases with the same ratio. Therefore, the scattering of the supervectors belonging to the negative class also increases. This increment of the scattering of the negative class data creates very high obstacle against the shifting of the decision hyperplane towards the negative class. Therefore, the claim made by Mak, M.W., & Rao,W., that the decision hyperplane shifts towards the negative class after application of the training utterance partitioning (UP) approach is not very logical. If more number of training supervectors in the positive class is useful to improve the recognition performance of the GMM-SVM classifier then, it should equally improve the recognition performance for test segments of short duration as well as for test segments of long duration. However, from the results it is very clear that, after application of the training utterance partitioning approach, the recognition performance of the GMM-SVM classifier has improved mainly for test segments of short duration (i.e., 5 s), for test segments of long duration (i.e., 20 s) the improvement of the recognition performance is very negligible (see Table 2 to Table 4). Therefore, we reject their claim of shifting the decision hyperplane towards the negative class due to the presence of more number of supervectors in the positive class after application of the training utterance partitioning approach.

In this experimental review and analysis article, we have postulated that, training utterance partitioning (UP) approach reduces the mismatch between the amounts of the MAP adaptations of training GMM supervectors and test GMM supervectors for the same speaker (see section 4, Fig.4 to Fig.6). Based on our postulation we have intuitively explained that, after application of the training utterance partitioning approach the scope for improvement of the recognition accuracy of the GMM-SVM classifier by reducing the numbers of false negatives is much more for test segments of short duration (i.e., 5 s) compared to the test segments of long duration (i.e., 20 s). Results of Table 2 to Table 4 and Fig.7 to Fig.9 strongly corroborate explanation based on our postulation. Therefore, based on the above discussions we firmly conclude the following:

> *Training utterance partitioning (UP) approach is not reducing the data imbalance problem of the GMM-SVM classifier. Although, we get more number of supervectors in the positive class after application of the UP approach, however, that is not the reason for the improvement of the recognition accuracy of the GMM-SVM classifier. After application of the UP approach, the recognition performance of the GMM-SVM classifier increases, because, the mismatch of the amounts of the MAP adaptations* *between training GMM supervectors and test GMM supervectors of the same speaker reduces which essentially reduces the numbers of false negatives.*

The intuitive explanation of conclusion (11) is given below:

It is evident from the results that, for test segments of long duration (i.e., 20 s), the speaker recognition performance of the GMM-SVM classifier is reasonably good when we partition the training utterance by two and construct two training GMM supervectors per speaker. It is also evident from the results that, for test segments of short duration (i.e., 5 s), the performance of the GMM-SVM classifier is reasonably good when we partition the training utterance by eight and construct eight training GMM supervectors per speaker. For a training utterance of duration 120 seconds, two and eight training GMM supervectors per speaker imply that, the duration of each sub-utterance used for preparation of training GMM supervector is 60 s and 15 s respectively. At this stage, the overall performances of the GMM-SVM classifiers saturate. After that, if we do more partitioning of the training utterance, the improvement of recognition performance for GMM-SVM classifier is negligible. Hence, we can say that, the duration of the test segments should not be less than one third of the duration of the training utterances.

# 6    EFFECT OF THE DIMENSION OF THE GMM SUPERVECTOR ON THE OVERLAP BETWEEN VARIOUS CLASSES IN THE GMM SUPERVECTOR DOMAIN

In this section, we investigate the effect of the dimension of the GMM supervector on the overlap between various classes. We select the average between-class-distance as the measure of overlap between various classes. If the average between-class-distance is more then it indicates that, the overlap between various classes is less. On the other hand, if the average between-class-distance is less then it indicates that, the overlap between various classes is more. The process of calculation for average between-class-distance is given below:

We calculate the Euclidean distance matrix **D** for *N* numbers of speakers. Each entry $d_{ij}$ of the distance matrix **D** represents the Euclidean distance between GMM supervector of class *i* and GMM supervector of class *j*. Therefore, the distance matrix **D** is a symmetric matrix and each entry of main diagonal is zero. The structure of the distance matrix **D** is given below:

$$\mathbf{D} = \begin{pmatrix} 0 & d_{12} & d_{13} & \cdots & d_{1N} \\ d_{12} & 0 & d_{23} & \cdots & d_{2N} \\ d_{13} & d_{23} & 0 & \cdots & d_{3N} \\ \vdots & \vdots & \vdots & & \vdots \\ d_{1N} & d_{2N} & d_{3N} & \cdots & 0 \end{pmatrix} \quad (9),$$

Where $d_{ij} = \|\text{GMM supervector of class } i - \text{GMM supervector of class } j\|_2$ (10)

Take the upper triangular portion of the distance matrix **D** and add all the $^{N}C_{2}$ numbers of terms. This represents the total distance (*TD*) between the classes as given below:

$$TD = \{d_{12} + d_{13} + d_{14} \cdots + d_{1N}\} + \{d_{23} + d_{24} + d_{25} \cdots + d_{2N}\} + \cdots + \{d_{(N-1)N}\}. \quad (11)$$

Therefore, average between-class-distance is given by the following equation:

$$\text{Average between-class-distance} = TD / (^{N}C_{2}). \quad (12)$$

Table 6 shows the average between-class-distance for three types of GMM supervector domains of dimensions 9728, 4864 and 2432.

Table 6 Average between-class-distance of all speakers for three types of GMM supervector domains.
SD= Supervector Dimension

| GMM Supervector Domain | Average between-class-distance |
|---|---|
| SD=9728 | 0.98 |
| SD=4864 | 1.23 |
| SD=2432 | 1.5 |

From the above results of average between-class-distance (see Table 6) we draw the following two important conclusions:

(12) *When the dimension of the GMM supervector domain increases, the overlap between various classes also increases.*

(13) *At the time of MAP adaptation, when we use UBM of relatively lower model order comparatively more numbers of Gaussians are affected.*

Conclusion (12) is obvious from the results of Table 6, because, when the dimension of the GMM supervector domain increases, the average between-class-distance decreases. Therefore, classes are less separable in the higher dimensional GMM supervector domain.

Conclusion (13) requires more detailed clarification. The intuitive explanation of conclusion (13) is given below:

Fig.11 shows a histogram from an artificially generated data set. It is evident from Fig. 11 that; the underlying distribution of the generated data set has three dominant modes.

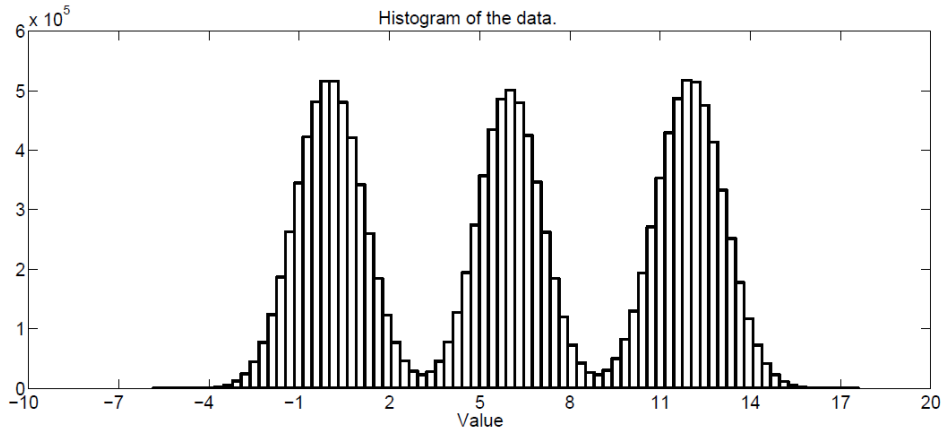

Fig.11 Histogram of the artificially generated data set.

Fig.12 shows the Gaussian mixture model of the above artificially generated data set. In this implementation, we have taken GMM model order three. It is clear from Fig.12 that, the estimated pdf (probability density function) closely follows the histogram of the data as shown in Fig.11. Fig.13 shows the Gaussian mixture model of the above artificially generated data set using GMM model order two. It is evident from Fig.13 that, the estimated pdf (probability density function) does not accurately follow the original histogram of the data as shown in Fig.11.

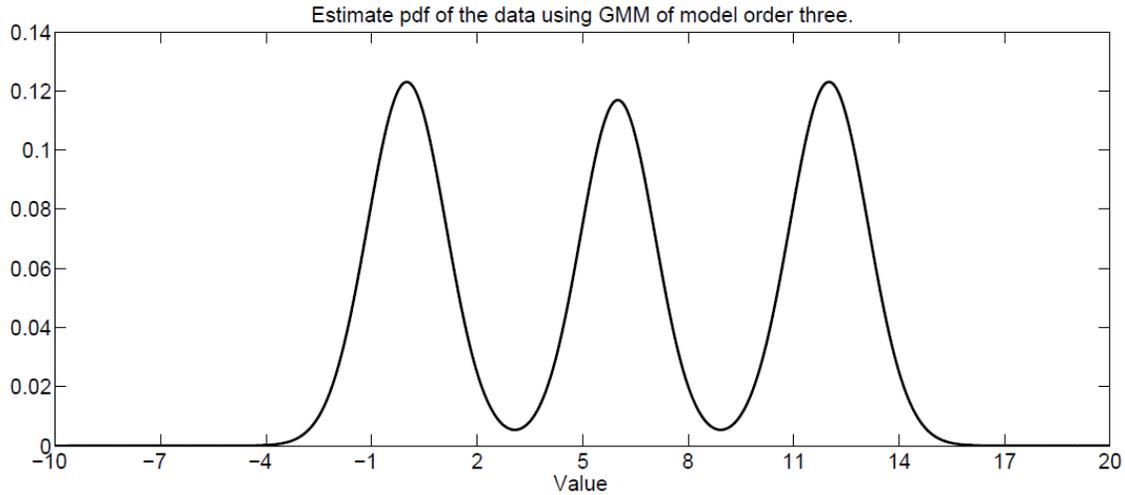

Fig.12 Approximate probability density function (pdf) of the above artificially generated data set using Gaussian mixture model with model order three.

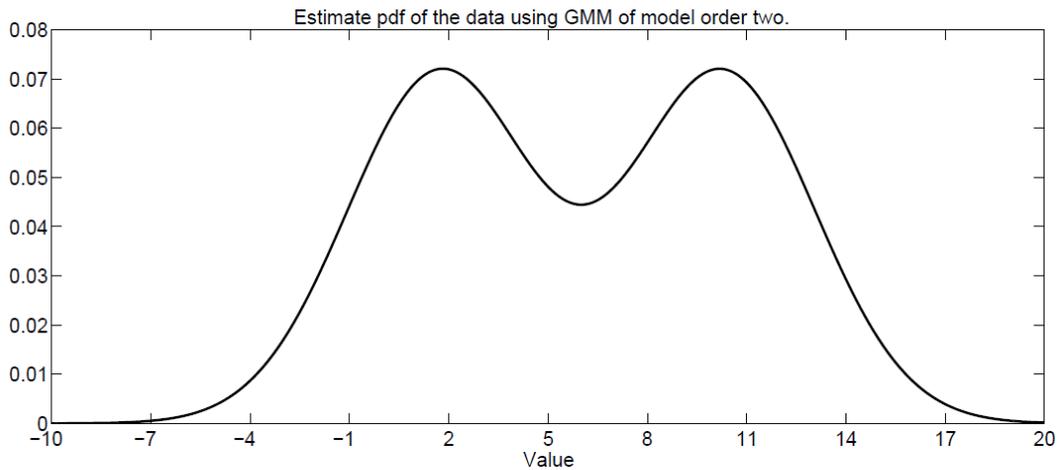

Fig.13 Approximate probability density function (pdf) of the above artificially generated data set using Gaussian mixture model with model order two.

Fig.14 and Fig.15 show individual Gaussian components of the Gaussian mixture model with model order three and model order two respectively. It is clear from Fig.14 and Fig.15 that, for lower-model-order representation, each individual Gaussian component has much larger variance compared to higher-model-order representation. Therefore, if we create the UBM with lower-model-order then at the time of MAP adaptation, training feature vectors affect higher number of Gaussians compared to the situation, when we prepare UBM with higher-model-order. For example, the training features which belong to the domain [-7, -4] do not affect any Gaussian in Fig.14. However, the same features affect one Gaussian in Fig.15. Similarly, training features which belong to the domain [4, 8] affect only one Gaussian in the Fig.14 but they affect two Gaussians simultaneously in Fig.15. Lastly, training features which belong to the domain [16, 19] do not affect any Gaussian in Fig.14. However, the same features affect one Gaussian in Fig.15. To keep the explanation simple, we have used single dimensional example. However, same explanation is valid for higher dimensional situation also.

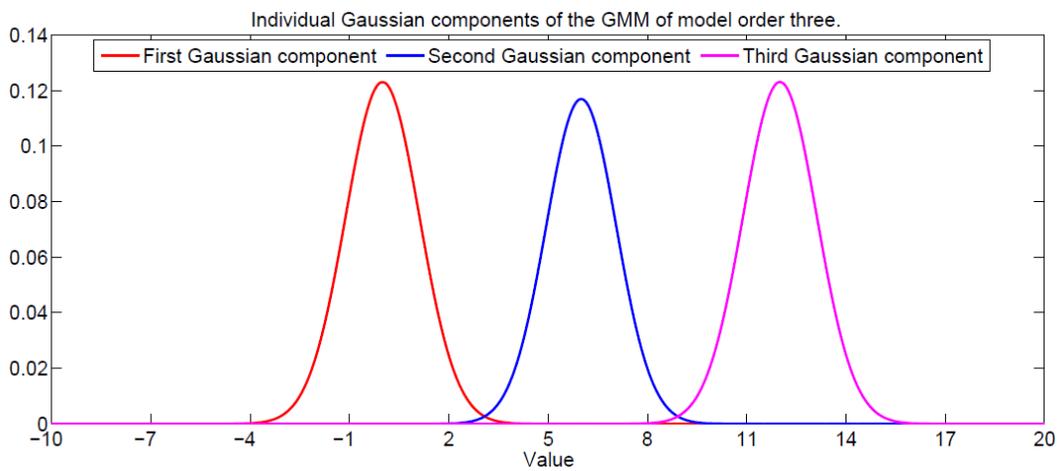

Fig.14 Individual Gaussian components of the Gaussian mixture model with model order three of the above artificially generated data set.

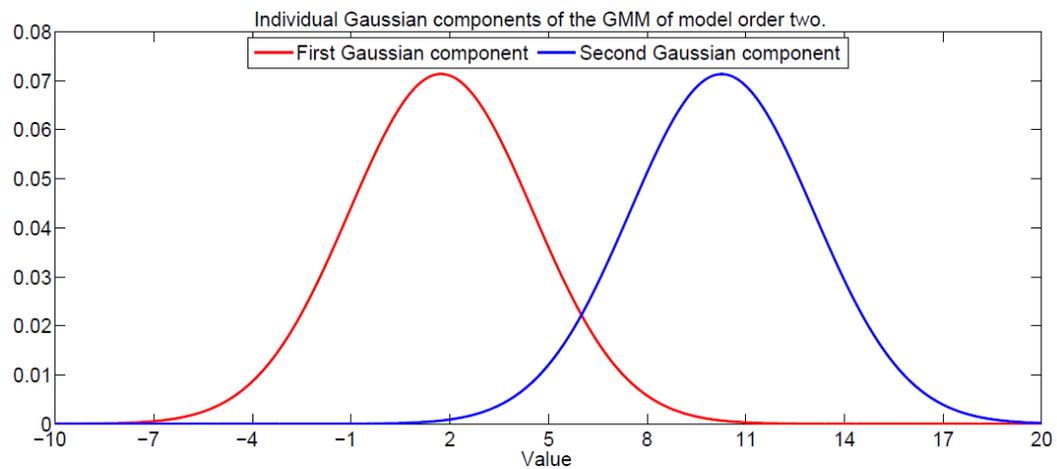

Fig.15 Individual Gaussian components of the Gaussian mixture model with model order two of the above artificially generated data set.

From the above discussion, it is clear that, when the model order of UBM is low then during MAP adaptation, more numbers of mean vectors of Gaussians of UBM change (because the variance of each Gaussian is large). However, when the model order of UBM is high then during MAP adaptation, less numbers of mean vectors of Gaussians of UBM change (because the variance of each Gaussian is small). Therefore, if the dimension of the GMM supervector is high (i.e., the model order of UBM is high), less numbers of components of the GMM supervector are different from "UBM supervector" (i.e., if we prepare a supervector from UBM then we call that supervector as "UBM supervector"). However, if the dimension of the GMM supervector is low (i.e., the model order of UBM is low), large numbers of components of the GMM supervector are different from "UBM supervector". This is the inherent reason of our previous observation that, when the dimension of the GMM supervector increases, the classes become more overlapping. However, if the model order of UBM is very low then, the estimated pdf does not capture the distribution correctly and the recognition accuracy of the GMM-SVM classifier reduces. Therefore, for GMM-SVM classifier it is very important to select the proper model order of UBM (i.e., the model order of UBM should not be very high and at the same time, it should not be very low).

# 7   AVERAGE NUMBER OF SUPPORT VECTORS REQUIRED BY POSITIVE CLASS AND NEGATIVE CLASS OF GMM-SVM CLASSIFIER BEFORE AND AFTER PARTITIONING OF TRAINING UTTERANCE

In this section, we investigate the average number of support vectors required by positive class and negative class of GMM-SVM classifier (i.e., SVM classifier working in the GMM supervector domain). Table 7 shows the average number of support vectors required by SVM classifier for three types of GMM supervector domains (i.e., SD=9728, SD=4864 and SD=2432) before and after partitioning of training utterance.

Table 7  Average number of support vectors required by positive class and negative class of SVM classifier for three types of GMM supervector domains.
SD= Supervector dimension. Numbers inside the parenthesis indicate the number of support vectors required by the different classifiers.

| No of Partitions of the Training Utterance | Average Number of Support Vectors Required by Positive Class | | | Average Number of Support Vectors Required by Negative Class | | |
|---|---|---|---|---|---|---|
| | SD=9728 | SD=4864 | SD=2432 | SD=9728 | SD=4864 | SD=2432 |
| NO | 1 | 1 | 1 | 20 | 17 | 15 |
| 2 | 2 | 2 | 2 | 30 | 25 | 21 |
| 3 | 3 | 3 | 3 | 41 | 31 | 26 |
| 4 | 4 | 4 | 3.96  (3-4) | 51 | 37 | 30 |
| 6 | 5.97  (5-6) | 5.92  (5-6) | 5.79  (4-6) | 75 | 51 | 38 |
| 8 | 7.86  (6-8) | 7.81  (6-8) | 7.51  (6-8) | 101 | 65 | 46 |
| 12 | 11.5 (8-12) | 11  (7-12) | 10.64 (7-12) | 166 | 96 | 64 |

It is evident from Table 7 that, when the dimension of the GMM supervector increases, the average number of support vectors required for training of SVM classifier also increases. We know from Section 6 that, when the dimension of the GMM supervector domain increases, the overlap between various classes also increases (conclusion 12). Based on the results in Table 7 and conclusion (12) we draw the following conclusion:

> (14) *When the overlap between positive class and negative class increases, more numbers of support vectors are required for SVM training.*

Conclusion (14) is very logical. The intuitive explanation for this conclusion is given below:

Fig.16 shows the SVM classifier with soft margin decision boundary [Burges, C. J. C., 1998; Alpaydin, E., 2004]. For a given training sets $(\mathbf{x}_i, y_i)$ of positive class and negative class (here, $\mathbf{x}_i$ is a training sample and $y_i$ is its class label), the decision boundary is explained by using the following equation:

$$\mathbf{w}^T\mathbf{x}_i + b \geq +1 - \xi_i \text{ for } y_i = +1 \quad \text{and} \quad \mathbf{w}^T\mathbf{x}_i + b \leq -1 + \xi_i \text{ for } y_i = -1 \quad (13)$$

Where **w** is the weight vector and $b$ is the bias. Slack variables, $\xi_i \geq 0$ are defined to measure the deviation from the margin. Deviations are of two types: a training sample may occur on the correct side of the decision boundary but in-between the decision boundary and the margin. This sample is classified correctly but with low confidence. On the other hand, a training sample may occur on the wrong side of the decision boundary and is misclassified. It is evident from the equation of the decision boundary that, for an error to occur, the corresponding $\xi_i$ must exceed unity. Fig.16 shows different positions possible for the support vectors of the SVM classifier with a soft margin decision boundary:

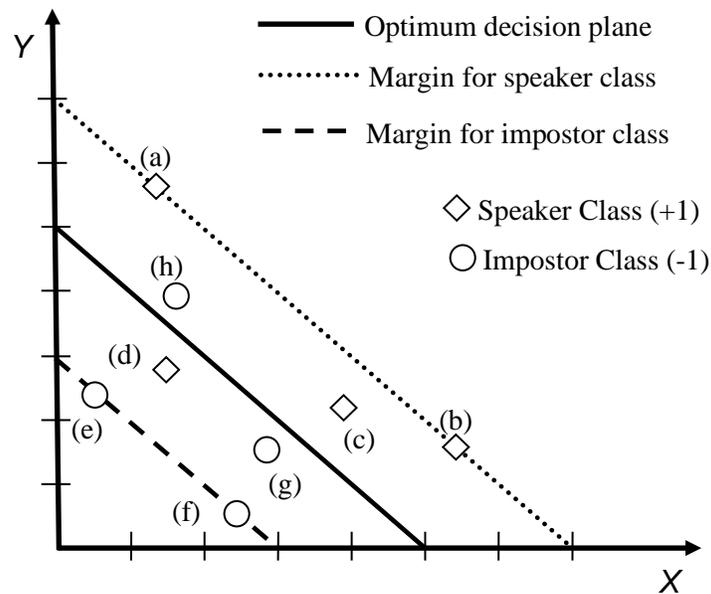

Fig.16 Different positions possible for the support vectors of the SVM classifier with a soft margin decision boundary.

Training samples (a), (b), (e) and (f) are on the correct side of the decision boundary and they are on their respective margins. Training samples (c) and (g) are on the correct side of the decision boundary but they occur in-between the decision boundary and their respective margins. Training samples (c) and (g) are classified correctly but with low confidence. Training samples (d) and (h) are on the wrong side of the decision boundary and they are misclassified. Therefore, the values of slack variables associated with above eight training samples are given below:

$$\xi_a = 0,\ \xi_b = 0,\ \xi_e = 0,\ \xi_f = 0;\ 0 < \xi_c < 1,\ 0 < \xi_g < 1;\ \xi_d > 1,\ \xi_h > 1 \qquad (14)$$

All the above eight training samples (i.e., training sample (a) to training sample (h)) are the support vectors for the SVM classifier with a soft margin decision boundary. Here, the support vectors (c), (d), (g) and (h) are overlapping. In case of SVM classifier with a soft margin decision boundary, the training samples which occur on the wrong side of the decision boundary are declared as the support vectors along with the training samples which occur on the margins and in-between the decision boundary and the margins. When the overlap between positive class and negative class increases, more numbers of training samples occur on the wrong side of the decision boundary. Therefore, the total numbers of support vectors increase. On the other hand, it is also very logical to think that, if the total numbers of support vectors increase, the numbers of support vectors, which are overlapping, also increase. In the extreme case, if all the support vectors are on the correct sides (i.e., $0 \leq \xi_i < 1$ for all $i$), then also, if the total numbers of support vectors increase, it implies that, more numbers of training samples from the two classes are close to each other.

Now, we are in a position to explain the conclusion (5) of this experimental review and analysis article (see section 4). The conclusion (5) reveals the relation between the dimension of the GMM supervector and the recognition performance of the GMM-SVM classifier. It is based on the observation that, when we do not apply the training utterance partitioning approach, the recognition performance of the GMM-SVM classifier for test segments of short duration (i.e., 5s) decreases when the dimension of the GMM supervector increases (see Fig.3). The intuitive explanation for conclusion (5) is given below:

We have already explained that, when we do not apply the training utterance partitioning approach, there is a large difference between the amounts of MAP adaptations of training GMM supervector and test GMM supervector for the same hypothesized speaker for test segments of short duration (see intuitive explanation of conclusions (2) and (3) of section 4). Therefore, at the time of testing, the false rejection rate is very high and the speaker recognition performance of the GMM-SVM classifier is very poor for test segments of short duration (i.e., 5s). From conclusion (12) of section 6 we know that, when the dimension of the GMM supervector domain increases, the overlap between various classes also increases. As a result, the accuracy for short duration test segments decreases even further when the dimension of the GMM supervector increases. Therefore, we conclude that, the greater extent of overlap between speaker classes in the higher dimension is the cause for the poor performance of GMM-SVM classifier for recognition of test segments of short duration in the higher dimension.

It is evident from the results of Table 7 that, when the dimension of the GMM supervector increases, the average number of support vectors required for training of SVM classifier also increases. Vapnik has given a bound on the expected test error rate for SVM classifier with respect to its numbers of support vectors as follows [Vapnik, V., 1995; Alpaydin, E., 2004]:

$$E[P(error)] \leq \frac{E[\text{Number of support vectors}]}{\text{Number of training samples}}, \quad (15)$$

Where $E[P(error)]$ is the expectation of the actual risk over all choices of training sets [Burges, C. J. C., 1998]. Hence, intuitively the SVM classifier which requires fewer numbers of support vectors with respect to the numbers of training samples is likely to give better performance. From the results of Table 7 it is clear that, SVM classifier which uses GMM supervectors of dimension 9728 requires many more support vectors compared to SVM classifier which uses GMM supervectors of dimension 4864 or 2432 for equal number of training samples. Hence, the observed phenomenon (i.e., when the dimension of the GMM supervector increases, the performance of the GMM-SVM classifier decreases) corroborates the error bound given by Vapnik. Therefore, conclusion (5) is logical. However, we understand that, if the dimension of the GMM supervector is very low (i.e., the model order of UBM is very low) then, the estimated pdf does not capture the distribution properly and the recognition accuracy of the GMM-SVM classifier reduces. Therefore, conclusion (5) essentially reveals that, we should choose the dimension of the GMM supervector very judiciously.

# 8    CONCLUSIONS

This experimental review and analysis article provides various conclusions with detailed intuitive explanations. In this section, we provide a comprehensive summary of those conclusions.

(1)    When we do not apply the training utterance partitioning approach, the recognition performance of GMM-SVM classifier for test segments of long duration is better than the classical GMM-UBM classifier. This is in accordance with the previous studies which shows that GMM-SVM system shows better performance than GMM-UBM classifier for test segments of long duration.

(2)    When we do not apply the training utterance partitioning approach, the recognition performance of GMM-SVM classifier for test segments of short duration is very poor compared to the classical GMM-UBM classifier. Therefore, for test segments of short duration, it is not feasible to use GMM-SVM classifier in a usual configuration.

(3)    When we have adequate amount of training data and we do not apply the training utterance partitioning approach for training the GMM-SVM classifier then, the degradation in recognition performance due to the reduction of duration of test segment is much less in classical GMM-UBM classifier compared to the GMM-SVM classifier.

(4)    In case of speaker recognition with GMM-UBM classifier, we first separately compute the score for each frame of the test utterance. Next, we find the mean of these scores. This "mean-score" represents the overall score of the test utterance. In this experimental review and analysis article, we have shown that, when the speaker models are trained with adequate amount of training data, the operation "mean" largely cancels out the effects of the duration of the test utterance. Hence, for a well-trained speaker model, the "mean-score" generated for a test segment of long duration and the "mean-score" generated for a test segment of short duration have less difference. Therefore, when the duration of training speech is adequate the degradation in recognition performance due to the reduction of duration of test segment is much less in classical GMM-UBM classifier.

(5) In this experimental review and analysis article, we have concluded that, when we do not apply the training utterance partitioning approach then, the poor performance of GMM-SVM classifier for recognition of test segments of short duration is not due to the data imbalance problem of the GMM-SVM classifier. Our conclusion is based on the facts that, data imbalance problem is completely related with the training condition of the classifier. Therefore, if the data imbalance problem is the cause for the poor recognition performance of the GMM-SVM classifier for the test segments of short duration then it should also affect the recognition performance of the GMM-SVM classifier for test segments of long duration. However, in this article we have shown that the recognition performance of the GMM-SVM classifier does not get adversely affected for test segments of long duration. Therefore, we discard the data imbalance problem as the probable cause for the poor performance of the GMM-SVM classifier for recognition of test segments of short duration. Hence, our conclusion essentially rejects the earlier hypothesis which was proposed by Mak, M.W., & Rao,W., in 2011 to explain the reason for the poor recognition performance of GMM-SVM classifier when we do not apply the training utterance partitioning approach, because, according to their proposition, data imbalance problem is the cause for the poor recognition performance of the GMM-SVM classifier.

(6) In this experimental review and analysis article, we have concluded that, when we do not apply the training utterance partitioning approach, due to the large mismatch of the amounts of MAP adaptations between training GMM supervector and test GMM supervector for the same hypothesized speaker, the GMM-SVM classifier performs very poorly for recognition of test segments of short duration. Our conclusion is based on the facts that, when we do not apply the training utterance partitioning approach and the duration of test speech is much less compared to the duration of training speech then, the numbers of test feature vectors are also much less compared to the numbers of training feature vectors. Hence, we prepare the training GMM supervector with the help of MAP adaptation technique of UBM by using the large numbers of training feature vectors. However, we prepare the test GMM supervector with the help of MAP adaptation technique of UBM by using the much lesser numbers of test feature vectors. Therefore, there is a large difference between the amounts of MAP adaptations of training GMM supervector and test GMM supervector for the same hypothesized speaker. Hence, at the time of testing, the false rejection rate is very high and the performance of GMM-SVM classifier for recognition of test segments of short duration is very poor. On the other hand, for test segments of long duration, the problem of mismatch of the amounts of MAP adaptations between training GMM supervector and test GMM supervector for the same hypothesized speaker is lesser prominent, because, in this case, the numbers of test feature vectors are not very less compared to the numbers of training feature vectors. Consequently, the recognition performance of the GMM-SVM classifier for test segments of long duration is satisfactory.

(7) After the application of training utterance partitioning (UP) approach, the recognition performance of GMM-SVM classifier for test segments of long duration has improved marginally.

(8) After the application of training utterance partitioning (UP) approach, the recognition performance of GMM-SVM classifier for test segments of short duration has improved very significantly.

(9) In this experimental review and analysis article, we have shown that, the mismatch of the amounts of MAP adaptations between training GMM supervector and test GMM supervector for the same speaker is much less when we apply the UP approach for training compared to the situation when we do not apply the UP approach for training. Consequently, numbers of false negatives, which are due to the mismatch of the amounts of MAP adaptations between training

GMM supervector and test GMM supervector for the same speaker are also much less when we apply the UP approach for training of the GMM-SVM classifier. Therefore, the recognition performance of the GMM-SVM classifier is better when we apply the UP approach for training compared to the situation if we do not apply the UP approach for training. However, in this experimental review and analysis article, we have explained in detailed that; the scope for improvement of the recognition performance of the GMM-SVM classifier, by reducing the mismatch of the amounts of MAP adaptations between training GMM supervector and test GMM supervector of the same speaker after the application of UP approach is much more for the test segments of short duration compared to the test segments of long duration. Therefore, after the application of training utterance partitioning (UP) approach, the performance of GMM-SVM classifier for recognition of test segments of short duration has improved very significantly, because, the scope for reduction of false negatives is very high for test segments of short duration. However, the performance of GMM-SVM classifier for recognition of test segments of long duration has improved marginally, because, the scope for reduction of false negatives is much less for test segments of long duration.

(10)    After we partitioned the training utterance by a suitable number, the GMM-SVM classifier performed far better than the classical GMM-UBM classifier even for recognition of test segments of short duration.

(11)    In this article, we have shown that, to achieve reasonably good recognition performance of the GMM-SVM classifier, the duration of the test speech segments should not be less than one third of the duration of the training speech utterance.

(12)    In this experimental review and analysis article, we have shown that, application of training utterance partitioning (UP) approach does not reduce the data imbalance problem of the GMM-SVM classifier. Our conclusion is based on the facts that, when we apply the training utterance partitioning (UP) approach for training of the GMM-SVM classifier, we apply UP for positive class as well as for negative class. Hence, positive class and negative class both get more numbers of supervectors. However, even after the application of UP approach also, the ratio between the numbers of supervectors of the positive class and the numbers of supervectors of the negative class remains constant (i.e., the ratio before application of UP and the ratio after application of UP is same). Hence, it is clear that, training utterance partitioning (UP) approach is not reducing the data imbalance problem of the GMM-SVM classifier. Consequently, we discard any possibility of reduction of data imbalance problem of the GMM-SVM classifier due to the application of UP approach. Therefore, we strongly reject the earlier hypothesis which was proposed by Mak, M.W., & Rao,W., in 2011, because, according to their proposition, after application of the training utterance partitioning approach, the performance of the GMM-SVM classifier improved due to the reduction of the data imbalance problem.

(13) In this experimental review and analysis article, we have practically discarded the possibility of shifting the decision hyperplane towards the negative class after the application of training utterance partitioning (UP) approach. Our conclusion is based on the facts that, when we apply the training utterance partitioning (UP) approach for training of the GMM-SVM classifier, positive class and negative class both get more numbers of supervectors. Consequently, scattering of the positive class data and scattering of the negative class data both increases. Increase of the scattering of the positive class supervectors after the application of UP approach tries to shift the decision hyperplane towards the negative class. However, the scattering of the supervectors belonging to the negative class also sufficiently increases, because, we have shown that, the application of UP approach does not reduce the data imbalance problem. This increment of the scattering of the negative class data creates very high obstacle against the shifting of the decision hyperplane towards the negative class. If at all, the decision hyperplane shifts towards the negative class then it should almost equally affect the recognition performance of the GMM-SVM classifier for test segments of long duration and for test segments of short duration, because, shifting of the decision hyperplane is entirely related to the training condition of the classifier. However, the recognition performance of the GMM-SVM classifier for test segments of short duration improved very significantly after the application of UP approach for training, but for test segments of long duration the recognition performance improved only marginally.

Hence, we virtually discard the possibility of shifting the decision hyperplane towards the negative class after the application of training utterance partitioning (UP) approach. Therefore, we reject the earlier hypothesis which was proposed by Mak, M.W., & Rao,W., in 2011, because, in their proposition, Mak & Rao postulated that, after application of the training utterance partitioning (UP) approach, the decision hyperplane shifts towards the negative class due to the presence of more numbers of supervectors in the positive class.

(14) In this experimental review and analysis article, we have shown that, when the dimension of the GMM supervector domain increases, the average between-class-distance decreases. Hence, overlap between various classes also increases. Therefore, we conclude that, classes are less separable in the higher dimensional GMM supervector domain.

(15) In this article, we have shown that, at the time of preparation of GMM supervectors by using the MAP adaptation from the universal background model (UBM), MAP adaptation process affects comparatively more numbers of Gaussians if we use UBM of comparatively lower model order.

(16) In this article, we have shown that, when we do not apply the training utterance partitioning approach then, the performance of GMM-SVM classifier for recognition of test segments of short duration is very poor. However, the recognition performance degrades further when the dimension of GMM supervector increases, because, classes are less separable in the higher dimensional GMM supervector domain. However, if the dimension of the GMM supervector is very low (i.e., the model order of UBM is very low) then, UBM does not estimate the distribution appropriately and the recognition accuracy of the GMM-SVM classifier reduces. Therefore, this experimental review and analysis article essentially says that, we should choose the dimension of the GMM supervector very judiciously.

(17) In this experimental review and analysis article, we have shown that, when the overlap between positive class and negative class increases, more numbers of support vectors are required for SVM training.

# ACKNOWLEDGEMENT

The authors are grateful to Dr. Goutam Saha, Department of E & ECE, IIT Kharagpur for his help in the experimentation with the POLYCOST database. First author is extremely grateful to Dr. Richa Mittal, erstwhile student of Department of CET, IIT Kharagpur for her help at the time of preparation of the manuscript. First author is also extremely grateful to Dr. Rahul Dasgupta, erstwhile student of Department of CET, IIT Kharagpur for rigorous technical discussions.